\documentclass[a4paper,twoside]{article}

\usepackage{epsfig}
\usepackage{subcaption}
\usepackage{calc}
\usepackage{amssymb}
\usepackage{amstext}
\usepackage{amsmath}
\usepackage{amsthm}
\usepackage{multicol}
\usepackage{pslatex}
\usepackage{apalike}
\usepackage{url}
\usepackage{xcolor}
\usepackage{algorithm2e}
\usepackage[bottom]{footmisc}
\usepackage{rotating}
\usepackage{booktabs}
\usepackage{graphicx} 

\usepackage{SCITEPRESS}     
\begin{document}

\title{SignIT: A Comprehensive Dataset and Multimodal Analysis for Italian Sign Language Recognition}

\author{\authorname{Alessia Micieli\sup{1}, Giovanni Maria Farinella\sup{1,2}, Francesco Ragusa\sup{1,2}}
\affiliation{\sup{1}LIVE@IPLab, Department of Mathematics and Computer Science - University of Catania, Italy}
\affiliation{\sup{2}Next Vision s.r.l., Spin-off of the University of Catania, Italy}
}


\abstract{In this work we present SignIT, a new dataset to study the task of Italian Sign Language (LIS) recognition. The dataset is composed of 644 videos covering 3.33 hours. We manually annotated videos considering a taxonomy of 94 distinct sign classes belonging to 5 macro-categories: Animals, Food, Colors, Emotions and Family. We also extracted 2D keypoints related to the hands, face and body of the users. With the dataset, we propose a benchmark for the sign recognition task, adopting several state-of-the-art models showing how temporal information, 2D keypoints and RGB frames can be influence the performance of these models. Results show the limitations of these models on this challenging LIS dataset. We release data and annotations at the following link: \url{https://fpv-iplab.github.io/SignIT/}.}

\onecolumn \maketitle \normalsize \setcounter{footnote}{0} \vfill

\begin{figure*}[t]
    \centering
    \includegraphics[width=\textwidth]{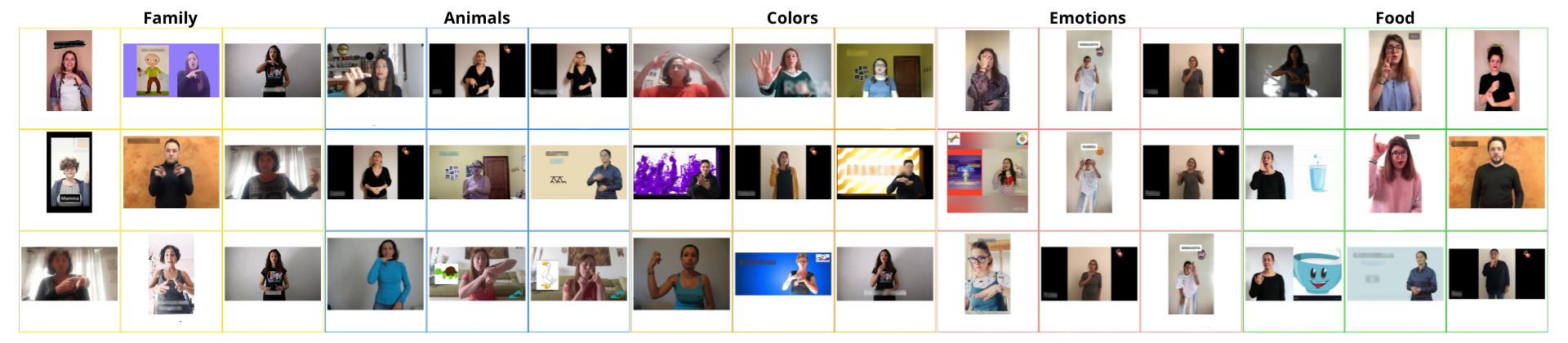}
    \caption{Samples of SignIT dataset across the 5 considered macro-categories: Animals, Food, Colors, Emotions, and Family.}
    \label{fig:grid}
\end{figure*}

\section{\uppercase{Introduction}}
\label{sec:introduction}

Sign languages represent a vital means of communication for deaf and hard-of-hearing individuals, allowing natural, expressive, and context-rich interaction within their communities and with hearing people. They play an essential role in reducing linguistic and social barriers, fostering inclusion in educational, professional, and everyday environments \cite{napier2009interpreting,marschark2002educating} .
Today, hundreds of distinct sign languages are used around the world [Book:\cite{brentari2010sign}], including American Sign Language (ASL), British Sign Language (BSL), and Chinese Sign Language (CSL), each with its own grammatical, lexical, and cultural characteristics.

Despite this diversity and social importance, communication between sign language users and those unfamiliar with sign languages remains a major challenge \cite{angelides2002accessibility,debevc2017sign}. This limitation underscores the need for advanced technologies capable of automatically recognizing and interpreting signs, facilitating more natural and inclusive interaction between deaf individuals and the broader hearing population.

Several studies have addressed the problem of sign language recognition across different languages, leveraging the availability of public datasets specifically designed for this task. In contrast, Italian Sign Language (LIS) remains relatively underexplored compared to more widely studied sign languages such as American Sign Language (ASL), British Sign Language (BSL), and Chinese Sign Language (CSL), primarily due to the lack of annotated datasets and standardized benchmarks \cite{mercurio2021lis2speech,schmalz2021realtime}.

In this work, we present a new dataset for Italian Sign Language (LIS), named SignIT, comprising 644 videos collected from publicly available online sources. The videos were recorded by \textcolor{red}{45} different users covering 3.33 hours. Videos have been manually annotated considering 94 distinct sign classes and organized into five macro-categories: \textit{Animals, Food, Colors, Emotions, and Family}. Figure~\ref{fig:grid} shows some examples of the SignIT dataset.
To enrich the dataset, we extracted pose keypoints at multiple levels of granularity, specifically for the face, hands, and body. To highlight the utility of the proposed dataset, we provide baseline results for the sign recognition task, focusing on three input modalities: (i) 2D human pose keypoints, (ii) visual appearance, and (iii) a combination of both.

\begin{figure}[!htbp]
    \centering
    \includegraphics[width=\linewidth]{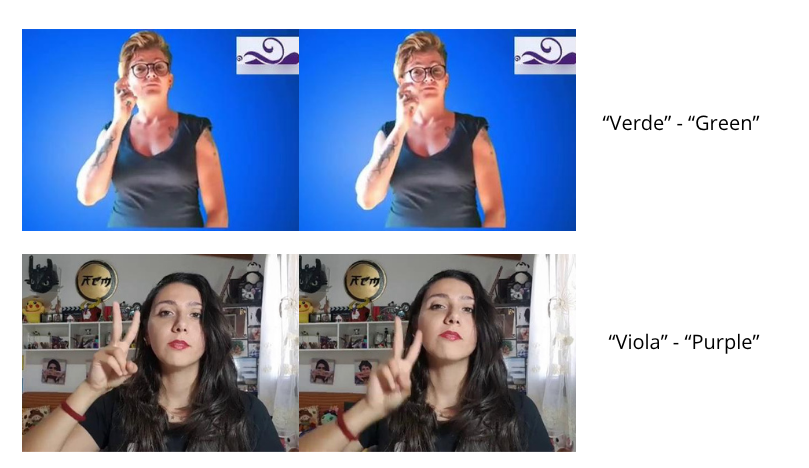}
    \caption{Example of similar signs in sign language.}
    \label{fig:similarity}
\end{figure}

The results indicate that current state-of-the-art approaches are insufficient to effectively address the LIS sign recognition task, particularly in cases where signs are visually similar and represent closely related concepts (e.g., green vs. purple).[Figure  \ref{fig:similarity}]

In sum the contributions of this work are as follows: 1) we present SignIT, a novel dataset for LIS recognition, consisting of 644 videos; 2) we manually annotated videos considering a taxonomy of 94 distinct sign classes; 3) we study the task of sign recognition exploiting 2D pose keypoints, visual appearance and a combination of both, showing the limitations of current state-of-the-art approaches. 
The dataset and its annotations are publicly available at: \url{https://fpv-iplab.github.io/SignIT/}.

\section{\uppercase{Related Work}}

\subsection{Sign Language Datasets}

\begin{table*}[!ht]
\centering
\caption{Comparison of recent approaches in sign language recognition.}
\label{tab:related_work}
\scriptsize
\setlength{\tabcolsep}{10pt}
\begin{tabular}{@{}l l l l c c r r@{}}
\toprule
\textbf{Dataset} & \textbf{Lang} & \textbf{Feat.} & \textbf{Task} & \textbf{Vid.} & \textbf{Year} & \textbf{Samples} & \textbf{Cls.} \\
\midrule
LIS Corpus~\cite{geraci2008grammar} & LIS & RGB & Linguistic corpus & V & 2010 & 16k & N/A \\
ISL-Ghotkar~\cite{isl_dataset} & ISL & RGB Hands & Basic signs &  & 2014 & 475 & 24 \\
JSL-Siamese~\cite{jsl_dataset} & JSL & RGB Hands & Fingerspell. &  & 2017 & 6k & 42 \\
PHOENIX14T~\cite{camgoz2020sign} & DGS & RGB/Pose & Sentences & V & 2018 & 7k & 108 \\
WLASL~\cite{lis2020word} & ASL & RGB & Word-level & V & 2020 & 21k & 2000 \\
CSL~\cite{huang2018video} & CSL & RGB/Depth & Alph.+2k & V & 2021 & 25k & 178 \\
BBC-Oxford~\cite{bsl_dataset} & BSL & RGB+Pose & Continuous & V & 2021 & 3.7M & 50k \\
Mediapi-RGB~\cite{fsl_dataset} & LSF & RGB+Text & Sentences & V & 2024 & 2.5M & 50k \\
\midrule
\textbf{SignIt (Ours)} & \textbf{LIS} & \textbf{RGB+Keypoints} & \textbf{Isolated gestures} & \textbf{V} & \textbf{2025} & \textbf{99k} & \textbf{94} \\
\bottomrule
\end{tabular}
\end{table*}

Extensive datasets have been developed for sign language recognition research across different languages. Multiple datasets exist for American Sign Language (ASL) \cite{lis2020word}, British Sign Language (BSL) \cite{albanie2020bsl}, Chinese Sign Language (CSL) \cite{joze2018ms}, and Filipino Sign Language (FSL) \cite{zhou2021improving}, enabling consistent benchmarking and model development. Conversely, Italian Sign Language (LIS) remains underrepresented, limiting opportunities for comparative analysis and robust recognition modeling. 

Existing LIS datasets are typically small-scale and focused on fingerspelling or limited vocabularies \cite{geraci2008grammar}. Our contribution addresses this gap with a larger dataset containing approximately 99,000 frames across 94 gesture classes, enriched with keypoints and RGB video data for systematic multimodal benchmarking.

Table~\ref{tab:related_work} provides a comprehensive comparison of existing state-of-the-art datasets.
While well-established resources exist for ASL, BSL, CSL, and other sign languages, Italian Sign Language (LIS) remains significantly underrepresented in the literature. Our dataset addresses this gap by offering a resource that integrates keypoints and RGB data across diverse semantic categories, providing the research community with a comprehensive benchmark for LIS recognition.

\textbf{ISL-Ghotkar}~\cite{isl_dataset} is a small-scale Indian Sign Language dataset containing 475 samples across 24 basic sign classes, using RGB hand-focused video recordings.

\textbf{JSL-Siamese}~\cite{jsl_dataset} focuses on Japanese Sign Language fingerspelling recognition with approximately 6,000 samples covering 42 characters, employing a Siamese network architecture for hand gesture recognition.

\textbf{LIS Corpus}~\cite{geraci2008grammar} is a linguistic corpus for Italian Sign Language (LIS) collected from 165 signers across 10 Italian cities, totaling approximately 165 hours of video data for linguistic analysis and grammar studies.

\textbf{PHOENIX14T}~\cite{camgoz2020sign} is a German Sign Language (DGS) dataset for continuous sign language translation, containing 7,000 sentence-level samples across 108 classes with both RGB video and pose annotations.

\textbf{WLASL}~\cite{lis2020word} (Word-Level American Sign Language) is a large-scale dataset with 21,000 video samples spanning 2,000 word-level signs, focusing on isolated word recognition in ASL.

\textbf{CSL}~\cite{huang2018video} is a Chinese Sign Language dataset featuring 25,000 samples with both RGB and depth modalities, covering the alphabet and approximately 178 sign classes.

\textbf{BBC-Oxford}~\cite{bsl_dataset} is a large-scale British Sign Language dataset containing 1,000 hours of broadcast footage with pose annotations, covering 50,000 vocabulary items for continuous sign language understanding.

\textbf{Mediapi-RGB}~\cite{fsl_dataset} is a recent French Sign Language dataset with 687 hours of annotated video and text transcriptions, encompassing 50,000 signs for sentence-level recognition tasks.

\subsection{Gesture Recognition Methods}

Sign language recognition has been addressed at different levels and through various methodological approaches.

Prior research has concentrated on alphabets and fingerspelling in sign languages such as ASL, ISL, and CSL, often using datasets composed of static postures \cite{wadhawan2021sign,rastgoo2021sign}. While valuable for isolated letter recognition, these approaches fail to address continuous or dynamic signing, limiting their applicability to real communication.  

Alternative methods rely on wearable or proximity sensors such as EMG, IMU, or UWB radar \cite{shin2023emg,sinyukov2016augmented,ahmed2023smart}. These systems can capture precise motion and physiological cues, enabling high recognition accuracy. However, they require specialized hardware, limiting scalability and usability in everyday settings. 

More challenging than alphabet or fingerspelling recognition is the task of full sign recognition, where the goal is to identify the semantic content conveyed through complex, dynamic hand movements \cite{koller2020sign,cui2019deep,camgoz2020sign}. These approaches achieve remarkable accuracy on isolated signs and complex gestures but depend heavily on large annotated datasets, high computational resources, and careful preprocessing. Recent trends focus on reducing annotation effort, improving temporal modeling, and enhancing generalization across signers and environments.

Recognition methods are strongly influenced by the dataset type and modality:

\textbf{Keypoint-based models:} Often implemented with lightweight classifiers, these methods are highly sensitive to occlusions, signer variability, and motion diversity~\cite{Alsharif2025,Ridwan2024,huang2018video,lis2020word}.
    
\textbf{CNN-based approaches:} Relying on raw image input, CNN methods require large datasets and significant computational resources, reducing applicability in real-time or resource-limited scenarios~\cite{Liang2018,Vashisth2023,cui2019deep,koller2019weakly}.

\textbf{Alphabet/fingerspelling-focused methods:} Many approaches still rely on small-scale or alphabet-only datasets, limiting generalization and hindering progress in LIS recognition~\cite{geraci2008grammar,marschark2002educating,schmalz2021realtime,akmeliawati2007real,wadhawan2021sign}.

In summary, while substantial progress has been made in sign language recognition through video-based and sensor-based methods, gaps remain in linguistic diversity, and LIS dataset availability.

\section{\uppercase{The SignIT Dataset}}

The SignIT dataset includes 94 distinct gesture classes belonging to the Italian Sign Language (LIS), distributed across five macro-categories: \textit{Animals, Food, Colors, Emotions}, and \textit{Family}. This semantic organization enables both category-level and fine-grained gesture analysis.

\subsection{Data Acquisition}

We collected a total of 644 videos from publicly available sources on the web, including user profiles on YouTube. The videos were originally organized into macro-categories such as animals, food, colors, emotions, and family, with roughly ten videos per category. Each video features a single user recording themselves with a frontal camera while performing multiple signs from Italian Sign Language (LIS).

To prepare the dataset for fine-grained analysis, each original video was manually segmented into individual sign instances, resulting in a total of 644 videos representing 94 distinct signs. All recordings took place in indoor environments across 37 different rooms, providing a variety of backgrounds and lighting conditions.

The SignIT dataset contains a total of 3 hours, 33 minutes, and 58.8 seconds of footage. Video durations average around 20 seconds, with resolutions ranging from 426×240 to 1024×1024 pixels, and frame rates between 24 and 30 FPS. This structure ensures that each video focuses on a single LIS sign, capturing detailed hand movements, facial expressions, and upper body gestures necessary for accurate recognition.

\subsubsection{Preprocessing}
\begin{figure}[!htbp]
    \centering
    \begin{minipage}[b]{0.48\textwidth}
        \centering
        \includegraphics[width=\textwidth]{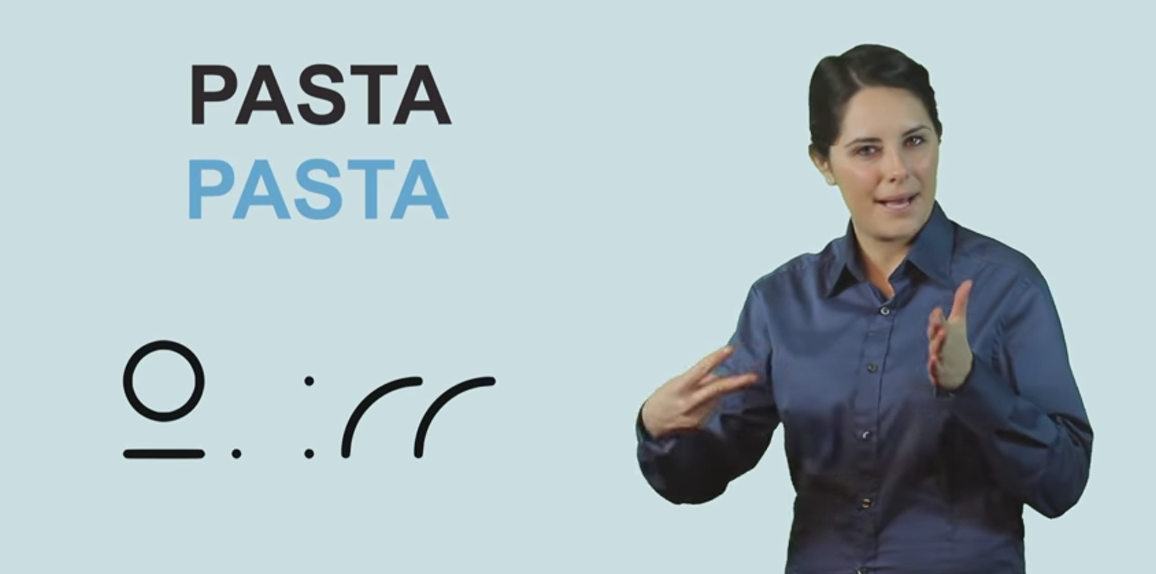} 
        \caption{Original Frame}
        \label{fig:originale}
    \end{minipage}
    \hfill
    \begin{minipage}[b]{0.48\textwidth}
        \centering
        \includegraphics[width=\textwidth]{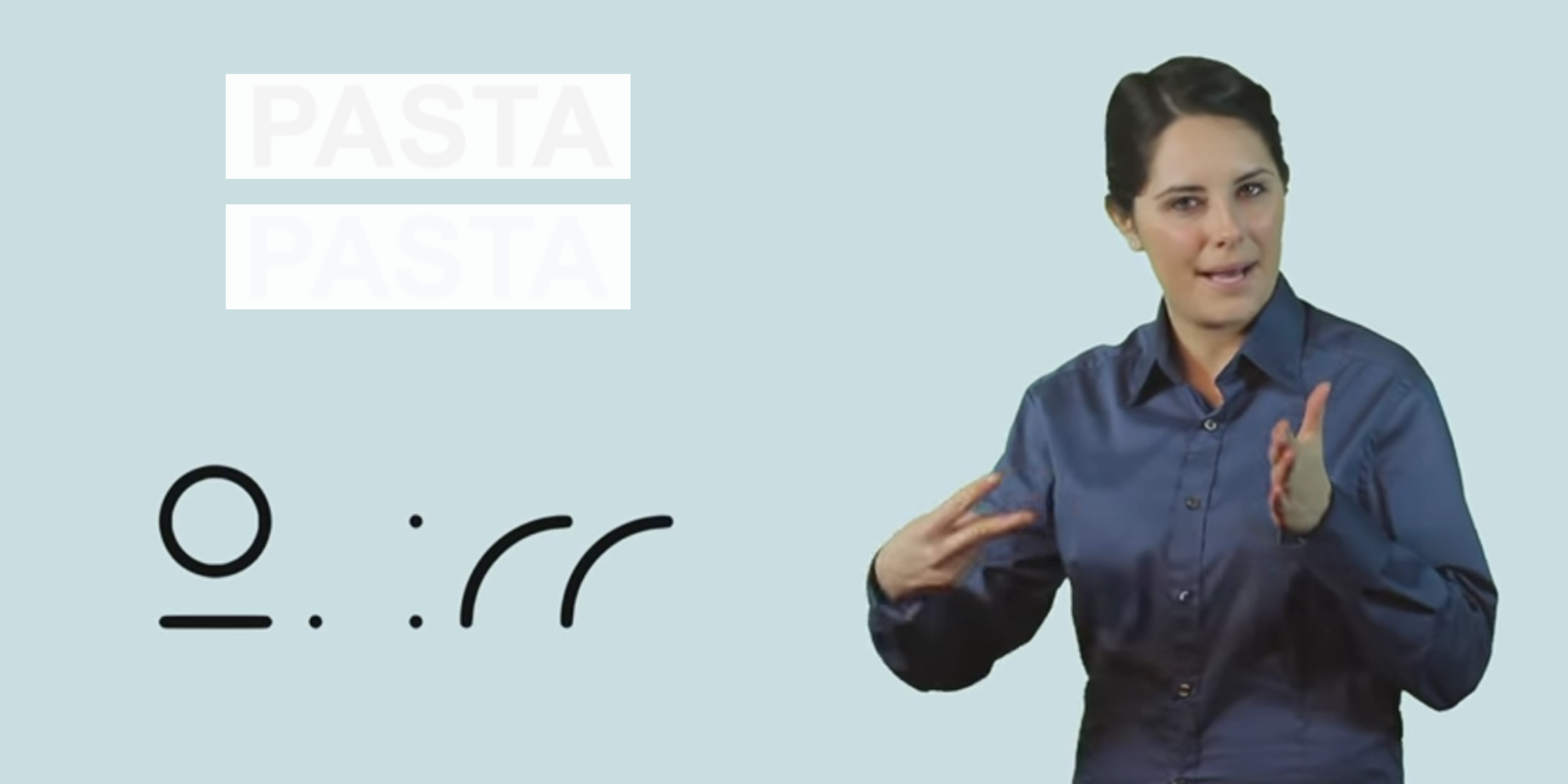} 
        \caption{Frame blurred}
        \label{fig:blurred}
    \end{minipage}
    \caption*{Comparison between the original image and the image after the application of the blurring process.}
\end{figure}

In several videos, textual or written content appears in the background, explicitly indicating the concept represented by the performed sign. To prevent recognition models from exploiting this information, rather than relying on the visual appearance or 2D keypoints of the signer, we applied a blurring operation to obscure these regions. Figure \ref{fig:blurred} shows the original frame (top) and the same frame after the preprocessing step (down).

\subsection{Data Annotation}
\begin{figure}[!htbp]
    \centering
    \includegraphics[width=0.9\linewidth]{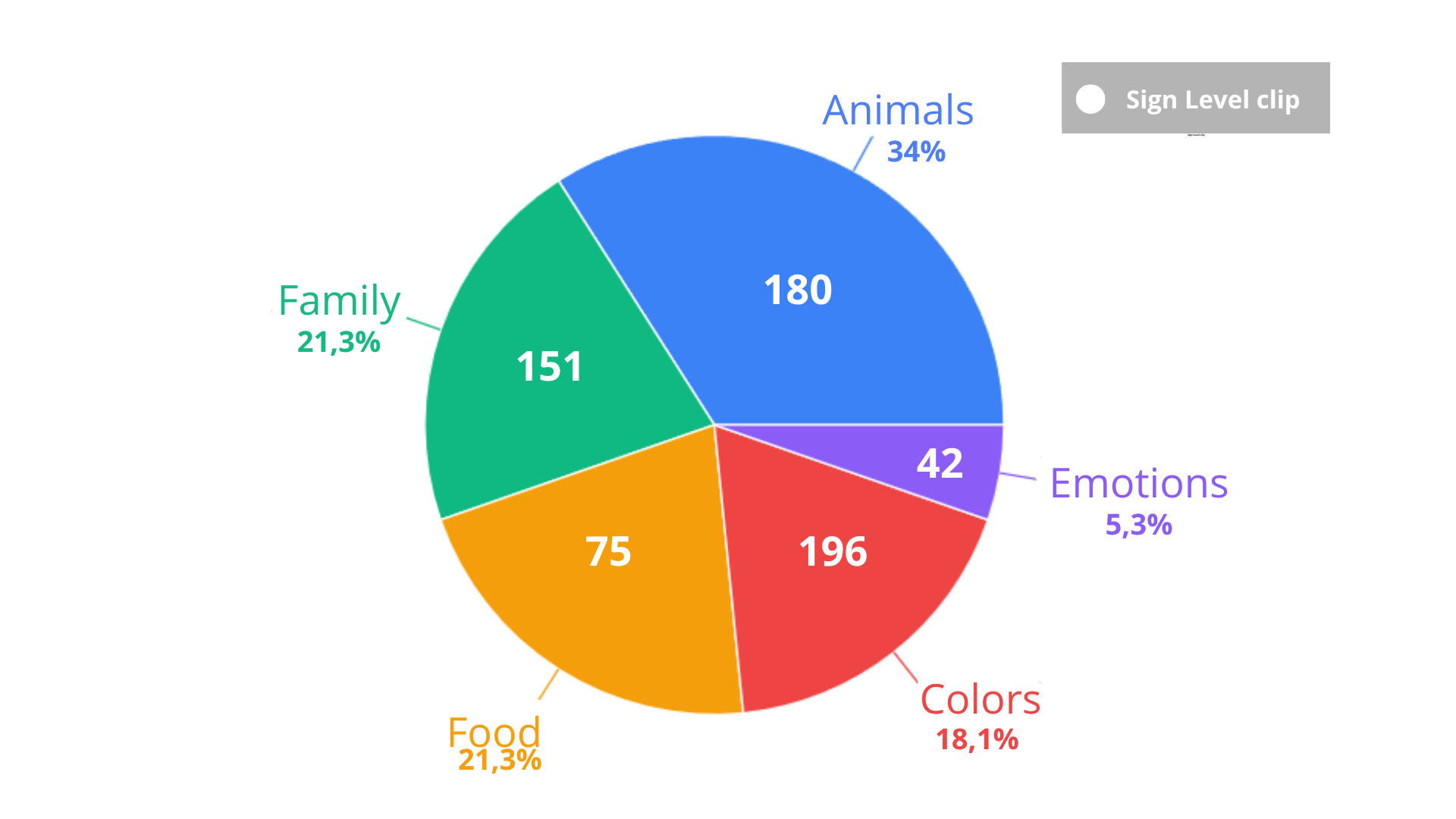}
    \caption{Macro-category distribution of the sign clips of the SignIT dataset.}
    \label{fig:statistic}
\end{figure}

We considered a taxonomy of 94 sign categories belonging to 5 macro-categories: Animals, Food, Colors, Emotions, and Family. We selected these 5 categories because they represent fundamental concepts that are typically among the first to be taught to individuals learning a sign language \cite{sign_for_child}. They encompass essential vocabulary that is commonly used in daily communication. 
These five macro-categories were specifically chosen because they form the foundational vocabulary for the most common words in sign language, aimed at teaching deaf children. The selection is intentional, as before learning complex sentences, deaf children are first taught individual words.
The videos were manually segmented into isolated signs, resulting in sign-level clips aligned with their corresponding gesture labels. Figure \ref{fig:statistic} shows the class distribution of SignIT along the 5 macro-categories.

\begin{figure*}[!htbp]
    \centering
    \begin{minipage}[b]{0.49\linewidth}
        \centering
        \includegraphics[width=\linewidth]{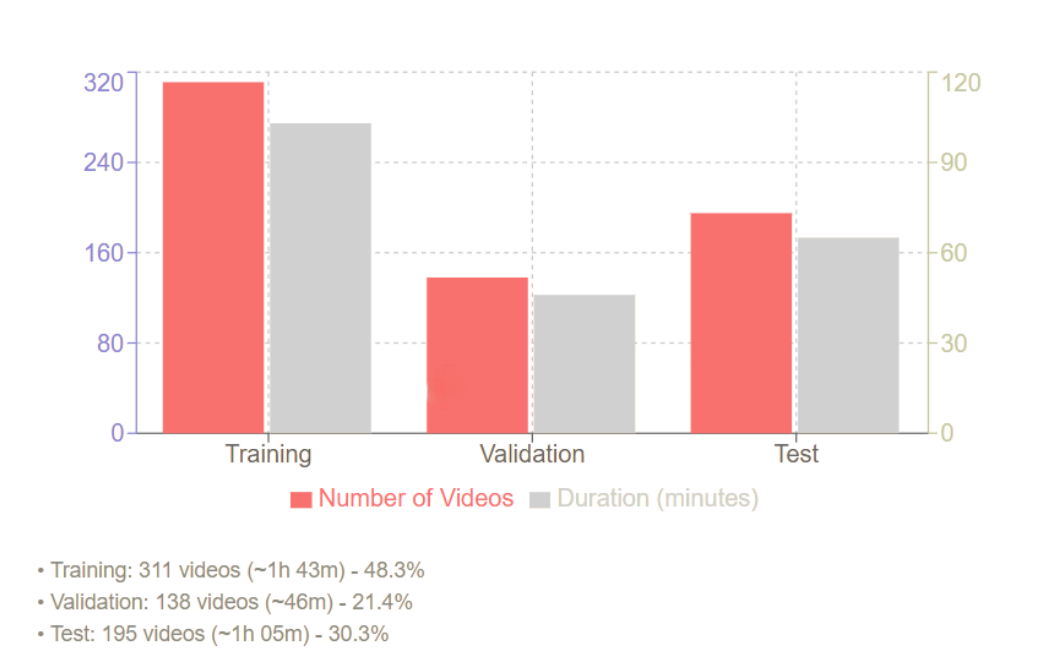}
        \caption{Total number of videos and the corresponding duration for each.}
        \label{fig:stat1}
    \end{minipage}
    \hfill
    \begin{minipage}[b]{0.49\linewidth}
        \centering
        \includegraphics[width=\linewidth]{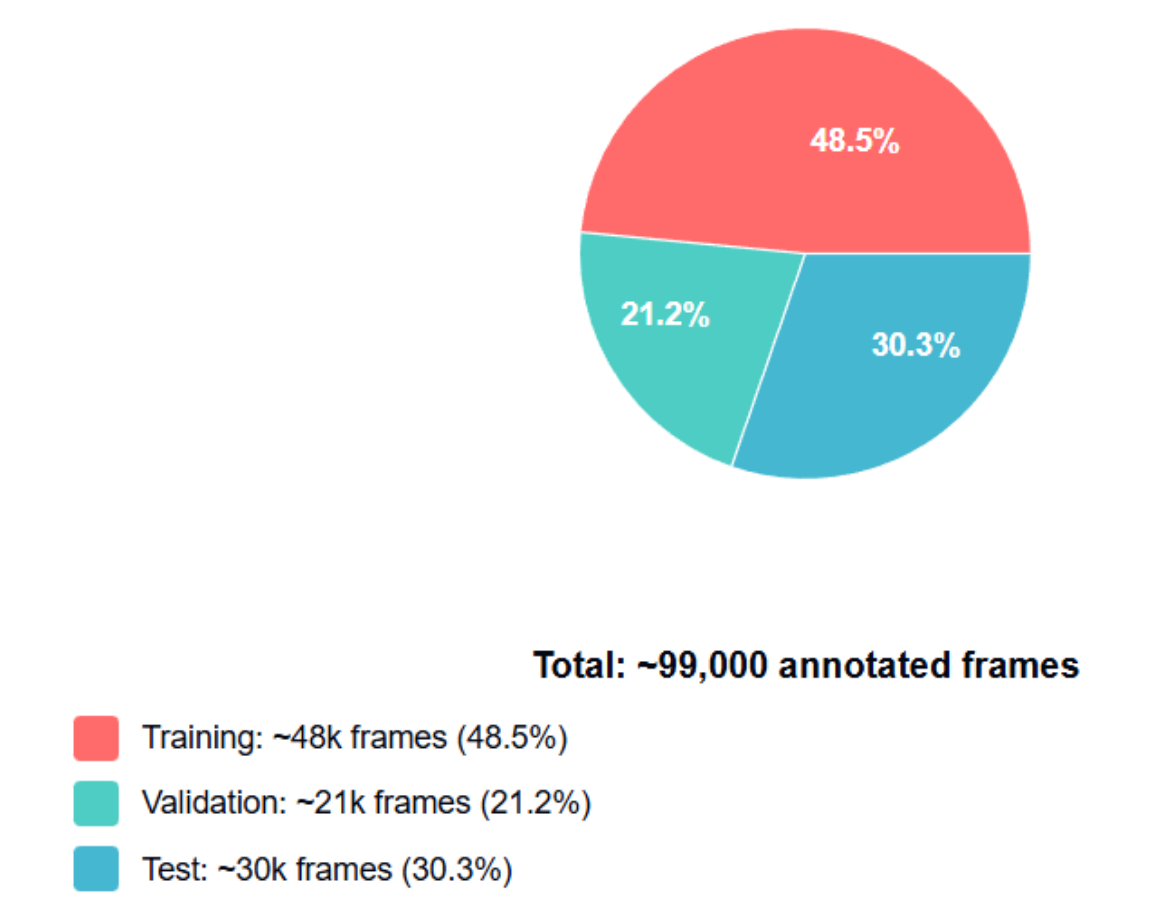}
        \caption{Distribution of samples across training, validation, and test sets.}
        \label{fig:stat2}
    \end{minipage}
\end{figure*}

The dataset was split into training (311 videos), validation (138 videos), and test (195 videos) subsets, ensuring a balanced distribution of gesture classes across all splits. Figure~\ref{fig:stat2} provides an overview of the dataset split, detailing both the number of videos and their total duration in each subset.

Specifically, the Training set comprises 311 videos with a total duration of approximately 1 hour and 43 minutes ($\sim$1h 43m). The Validation set includes 138 videos totaling about 46 minutes ($\sim$46m), and the Test set consists of 195 videos with a total duration of approximately 1 hour and 5 minutes ($\sim$1h 05m).

This distribution, which represents the number of videos, is consistent with the overall dataset proportions: training accounts for $\mathbf{48.5\%}$ of the total dataset, validation for $\mathbf{21.2\%}$, and test for $\mathbf{30.3\%}$. The distribution of annotated frames further confirms these proportions, as shown in the pie chart: $\sim$48k frames for training ($\mathbf{48.5\%}$), $\sim$21k for validation ($\mathbf{21.2\%}$), and $\sim$30k for testing ($\mathbf{30.3\%}$), resulting in a Total of $\sim$99,000 annotated frames. The bar chart visually highlights that the duration (in minutes) and the number of videos follow a similar, proportional trend across the three splits.

Figure~\ref{fig:statistic} further breaks down the class-level distribution within each category, revealing important intra-category imbalances. In the Animals category (Figure~\ref{fig:animals}), the top three classes (i.e., dog, turtle, horse) dominate with over 1,200 samples each, while the classes with the fewest examples include spider, monkey, and snake, each containing fewer than 200 samples.
This long-tail distribution poses a challenge, as some classes have very few examples, making it necessary to carefully handle class imbalance during training.
The Food category (Figure~\ref{fig:food}) exhibits a more balanced distribution, though prominent classes such as apple, banana, water still exceed 1,500 samples, while underrepresented classes like salt, rice, wine form the long tail. The Colors category (Figure~\ref{fig:colors}), with only 20 classes and a relatively uniform distribution, highlights high-frequency classes such as white, yellow, green, while minor classes like fuchsia, light, dark remain less represented. Emotions (Figure~\ref{fig:emotions}), with only five classes, has anger as the dominant class and disgust as the minor one. Family (Figure~\ref{fig:family}) categories display intermediate variability, with high-frequency signs such as dad, sister, grandfather and low-frequency ones including relatives, brother-in-law, daughter-in-law, reinforcing the importance of pose-based models that can leverage fine-grained spatial cues to distinguish between semantically similar gestures.

\subsubsection{2D Key-point Extraction}
\begin{figure}[!htbp]
    \centering
    \includegraphics[width=\linewidth]{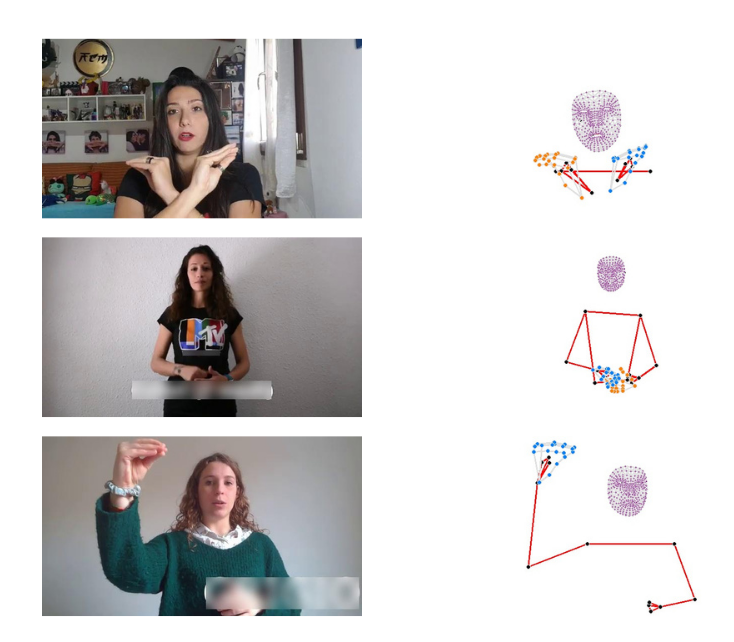}
    \caption{Examples of keypoint annotations for hands, face, and body belonging to the SignIT dataset.}
    \label{fig:keypoints}
\end{figure}

We extracted 2D keypoints using MediaPipe \footnote{\url{https://mediapipe.dev}} at different levels of granularity. For each frame, hand, face, and body keypoints were extracted. The hand model includes 21 keypoints per hand, enabling fine-grained tracking of finger articulation. The facial model includes 51 keypoints capturing the lips, nose, eyebrows, eyes, and facial contour, while the body model provides 33 keypoints representing upper-body posture and motion. Figure~\ref{fig:keypoints} illustrates examples of keypoint extractions from SignIT.

\section{\uppercase{Benchmark and Baseline Results}}

\subsection{Baselines}
To address the task of gesture recognition, we considered different baselines that are based on three input modalities: 1) 2D human pose keypoints, 2) visual appearance, and 3) a combination of both.
\\
\textbf{K-NN}: a standard K-Nearest Neighbors classifier that takes as input 2D keypoints corresponding to the hands, face, and body. These keypoints are concatenated into a single linear feature vector and standardized using Z-score normalization before classification.
\\
\textbf{MLP:} a standard Multi-Layer Perceptron composed of three fully connected layers, which takes the same input as the K-NN baseline.
\\
\textbf{ResNet18 \cite{He2016DeepResidual}:} a 2D networks which takes as input the whole RGB image to perform sign classification.
\\
\textbf{I3D \cite{Carreira2017I3D}:}
a 3D network which takes in input clips of 16 consecutive frames, each resized to $224$×$224$ pixels, sampled from the input video. 
\\
\textbf{LLaVA-OneVision~\cite{llavaonevision2024}:}
We adopted the LLaVA-OneVision-Qwen2-7B model to assess the capabilities of multimodal large language models (MLLMs) in recognizing signs from visual input. The model receives as input the RGB frame with Pose associate at frame and a prompt as follows:

\begin{quote}
\textit{Analyze this Italian Sign Language (LIS) video. The sign represents one concept from these categories: \\
1. animals \\
2. food \\
3. colors \\
4. family \\
5. emotions \\
Which category does this sign belong to? Respond with only: animals, food, colors, family, or emotions.}
\end{quote}

For macro-categories, a prompt was used to fine-tune and perform inference with LLaVA. This prompt provided general guidance on distinguishing broad sign categories based on key features, ensuring accurate classification at a higher level.

For micro-categories, a similar prompt was used, but adapted for each specific class due to the large number of classes, which required minor modifications to maintain precision.

\begin{figure*}[!htbp] 
    \centering
    \begin{subfigure}[b]{0.32\linewidth}
        \includegraphics[width=\linewidth]{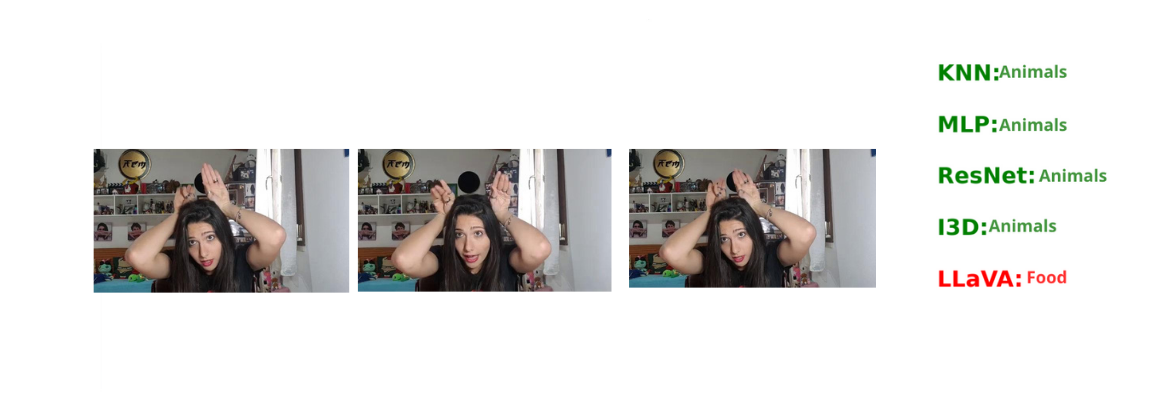}
        \caption{Animals}
        \label{fig:animals}
    \end{subfigure}
    \hfill
    \begin{subfigure}[b]{0.32\linewidth}
        \includegraphics[width=\linewidth]{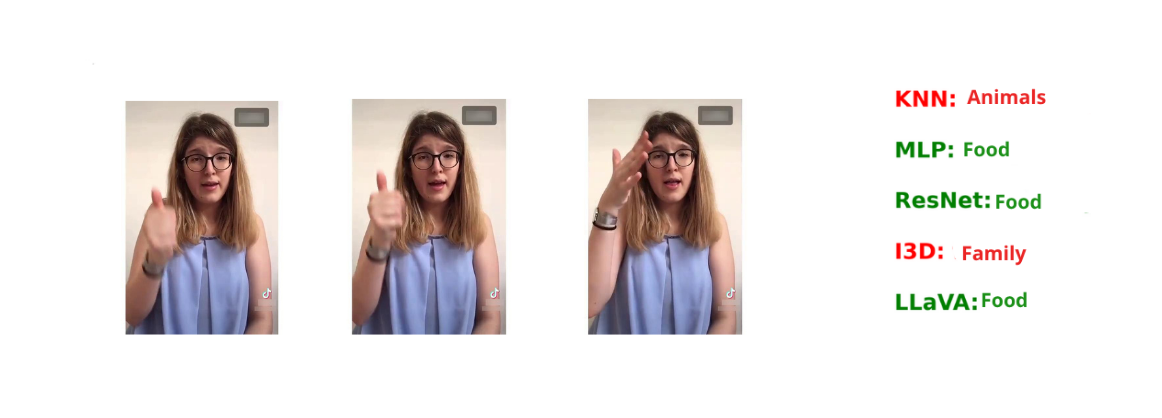}
        \caption{Food}
        \label{fig:food}
    \end{subfigure}
    \hfill
    \begin{subfigure}[b]{0.32\linewidth}
        \includegraphics[width=\linewidth]{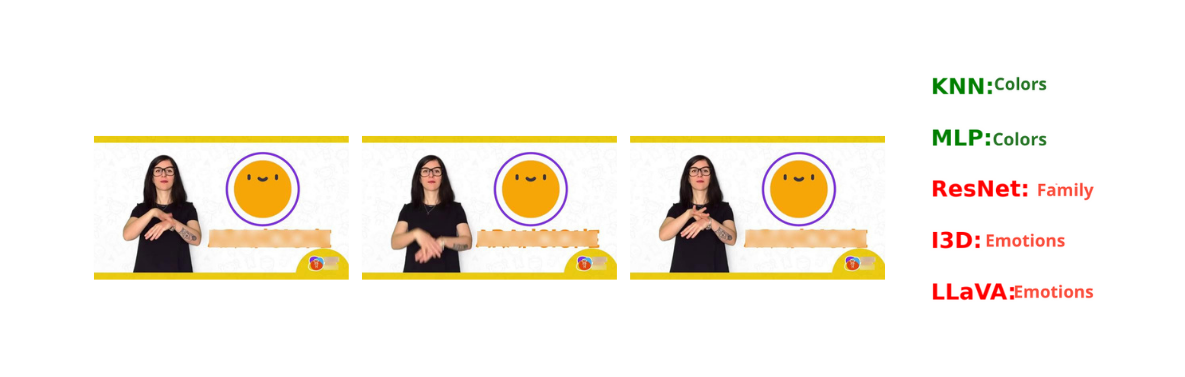}
        \caption{Colors}
        \label{fig:colors}
    \end{subfigure}

    \vskip 0.5em 

    \begin{subfigure}[b]{0.32\linewidth}
        \includegraphics[width=\linewidth]{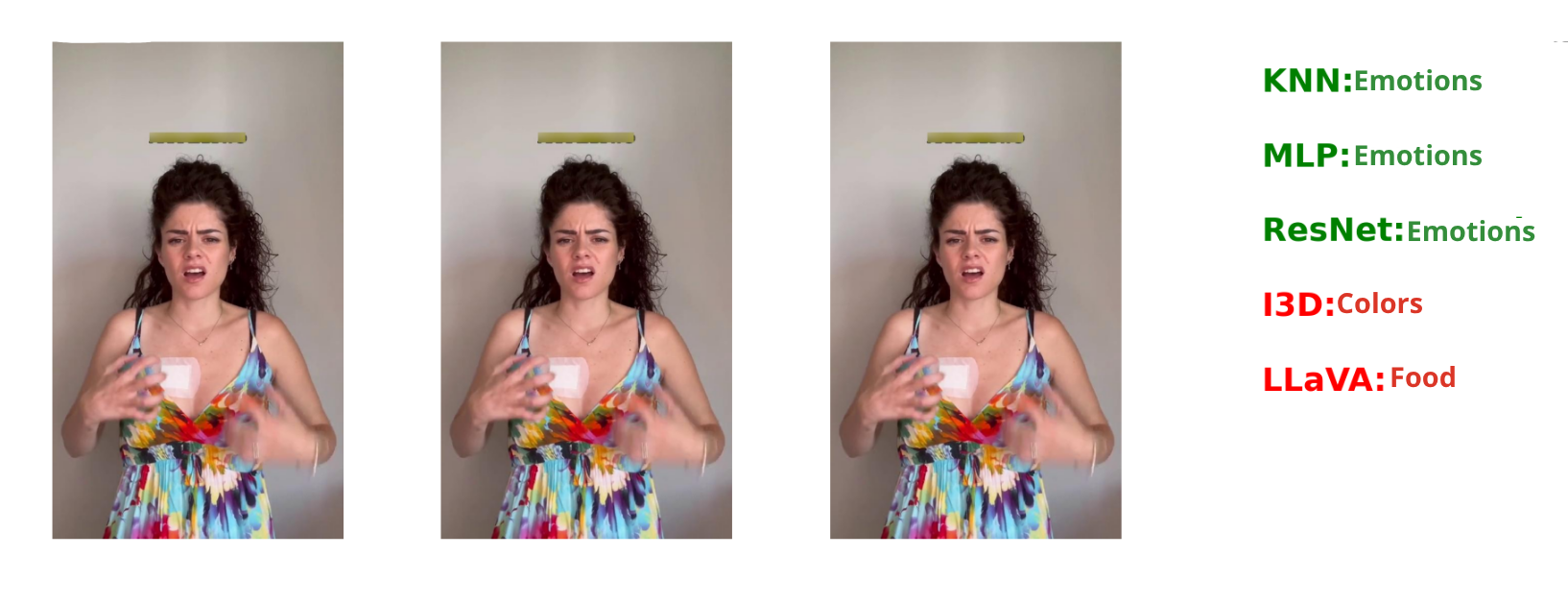}
        \caption{Emotions}
        \label{fig:emotions}
    \end{subfigure}
    \hspace{1em} 
    \begin{subfigure}[b]{0.32\linewidth}
        \includegraphics[width=\linewidth]{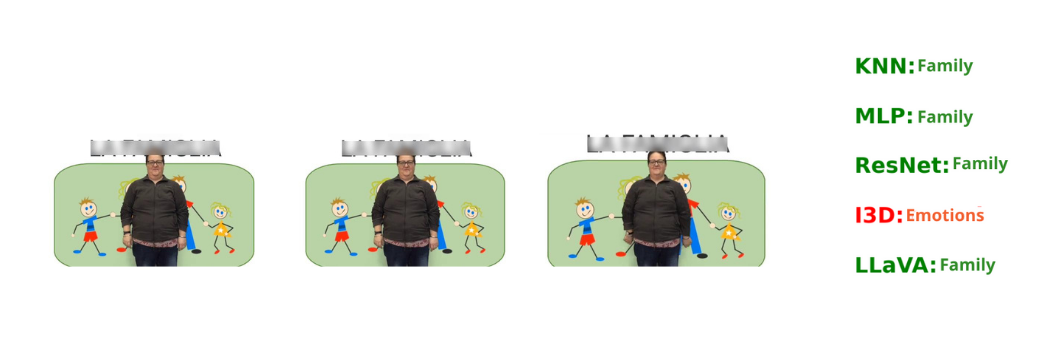}
        \caption{Family}
        \label{fig:family}
    \end{subfigure}

    \caption{Representative Examples for each macro-category of the SignIT dataset.}
    \label{fig:qualitative_results}
\end{figure*}

\subsection{Evaluation Measures}
We evaluate gesture recognition using standard accuracy computed on the whole test set. As class-aware measures, we report class-mean precision, recall and F1-score.

\subsection{Results}
This section presents a comprehensive evaluation of the adopted baselines across two recognition paradigms. We report performance metrics at both macro-category and fine-grained (micro) classification levels.

\begin{table*}[ht!]
\centering
\caption{Performance evaluation across models for macro-categories (overall) and individual categories of the SignIT dataset. Best results are shown in \textbf{bold}, second-best are \underline{underlined}. Acc = Accuracy, P = Precision, R = Recall.}
\label{tab:comprehensive_performance_corrected}
\resizebox{\textwidth}{!}{
\setlength{\tabcolsep}{2.5pt}
\begin{tabular}{@{}lcccc|cccc|cccc|cccc|cccc|cccc@{}}
\toprule
& \multicolumn{4}{c}{\textbf{Macro}} & \multicolumn{4}{c}{\textbf{Animals}} & \multicolumn{4}{c}{\textbf{Food}} & \multicolumn{4}{c}{\textbf{Colors}} & \multicolumn{4}{c}{\textbf{Emotions}} & \multicolumn{4}{c}{\textbf{Family}} \\
\cmidrule(lr){2-5} \cmidrule(lr){6-9} \cmidrule(lr){10-13} \cmidrule(lr){14-17} \cmidrule(lr){18-21} \cmidrule(lr){22-25}
\textbf{Model} & \textbf{Acc} & \textbf{F1} & \textbf{P} & \textbf{R} & \textbf{Acc} & \textbf{F1} & \textbf{P} & \textbf{R} & \textbf{Acc} & \textbf{F1} & \textbf{P} & \textbf{R} & \textbf{Acc} & \textbf{F1} & \textbf{P} & \textbf{R} & \textbf{Acc} & \textbf{F1} & \textbf{P} & \textbf{R} & \textbf{Acc} & \textbf{F1} & \textbf{P} & \textbf{R} \\
\midrule
KNN & 0.026 & 0.028 & 0.027 & 0.038 & 0.024 & 0.019 & 0.024 & 0.018 & 0.013 & 0.010 & 0.013 & 0.078 & 0.040 & 0.040 & 0.041 & 0.038 & 0.021 & 0.021 & 0.021 & 0.019 & 0.034 & 0.048 & 0.034 & 0.035 \\
MLP & \textbf{0.726} & \textbf{0.706} & \textbf{0.719} & \textbf{0.755} & \textbf{0.819} & \textbf{0.817} & \textbf{0.873} & \textbf{0.768} & \underline{0.635} & \underline{0.620} & \textbf{0.761} & 0.524 & \textbf{0.731} & \textbf{0.731} & \textbf{0.747} & \textbf{0.716} & 0.607 & 0.520 & 0.359 & \textbf{0.942} & \underline{0.840} & \textbf{0.840} & \textbf{0.854} & \textbf{0.827} \\
ResNet-18 & \underline{0.671} & 0.646 & 0.633 & \underline{0.675} & 0.655 & 0.635 & 0.616 & 0.654 & 0.627 & 0.591 & 0.489 & \underline{0.745} & \underline{0.646} & \underline{0.620} & \underline{0.701} & 0.556 & \underline{0.617} & \underline{0.597} & \underline{0.562} & \underline{0.637} & 0.812 & \underline{0.788} & \underline{0.796} & 0.781 \\
I3D & 0.386 & 0.340 & 0.347 & 0.304 & 0.196 & 0.267 & 0.500 & 0.182 & 0.309 & 0.235 & 0.200 & 0.286 & 0.231 & 0.316 & 0.600 & 0.214 & 0.309 & 0.400 & 0.666 & 0.286 & \textbf{0.884} & 0.462 & 0.321 & \underline{0.818} \\
LLaVA-OneVision & 0.656 & \underline{0.656} & \underline{0.648} & 0.645 & \underline{0.743} & \underline{0.730} & \underline{0.712} & \underline{0.750} & \textbf{0.721} & \textbf{0.706} & \underline{0.643} & \textbf{0.783} & 0.589 & 0.579 & 0.579 & \underline{0.579} & \textbf{0.672} & \textbf{0.659} & \textbf{0.707} & 0.617 & 0.557 & 0.545 & 0.600 & 0.500 \\
\bottomrule
\end{tabular}
}
\end{table*}

\subsubsection{Macro-categories}
Table \ref{tab:comprehensive_performance_corrected} reports the results obtained with the adopted baselines for the sign recognition task considering the 5 macro-categories of the SignIT dataset: \textit{Animals, Food, Colors, Emotions, and Family}. 
Overall, MLP achieved the best performance across all metrics (second row), demonstrating that the five macro-categories can be effectively discriminated using a 2D keypoints of hands, face and body. In terms of accuracy and recall, the second best result has been obtained by ResNet-18 (0.671 and 0.675), while the LlaVA-OneVision model (last row) ranked second considering the F1 score (0.656), and precision (0.648). Considering that F1 score is not influenced by the class unbalance of the SignIT dataset, since LLava-OneVision has been fine-tuned using frames and 2D keypoints, these results suggest that 2D keypoints carry strong discriminative information, contributing significantly to distinguishing between the five macro-categories. Figure \ref{fig:qualitative_results} reports some qualitative results.

Analyzing individual categories focusing on the F1 score, MLP  outperforms other baselines for \textit{Animals} (0.817), \textit{Colors} (0.731), and \textit{Family} (0.840), indicating that keypoint-based representations effectively help to recognize gestures belonging to these classes. LLaVA-OneVision achieves the best performance for \textit{Food} (0.706) and \textit{Emotions} (0.659), suggesting that combining visual frames with keypoints helps distinguish signs related to these classes.

\begin{table}[!htbp]
\centering
\caption{Comparison of LLaVA inference with and without pose information.}
\label{tab:pose_comparison}
\scriptsize
\setlength{\tabcolsep}{3pt}
\renewcommand{\arraystretch}{1.2}
\begin{tabular}{@{}p{0.30\columnwidth}p{0.15\columnwidth}p{0.48\columnwidth}@{}}
\toprule
\textbf{Frame} & \textbf{Inf.} & \textbf{GT / Predictions} \\
\midrule
\includegraphics[width=0.40\columnwidth,keepaspectratio]{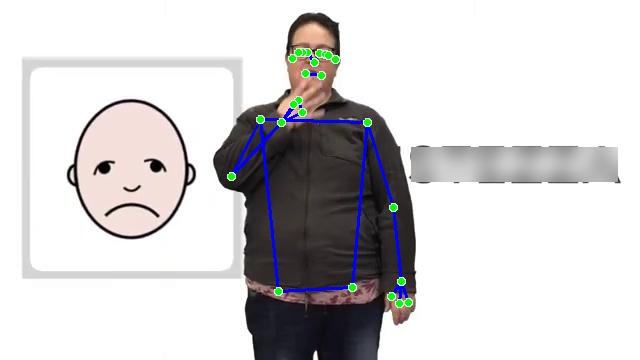} & 
\centering True & 
\begin{tabular}{@{}l@{}}
GT: \textit{emotion\_sadness} \\
Pred: \textit{emotion\_sadness}
\end{tabular} \\[6pt]
\includegraphics[width=0.40\columnwidth,keepaspectratio]{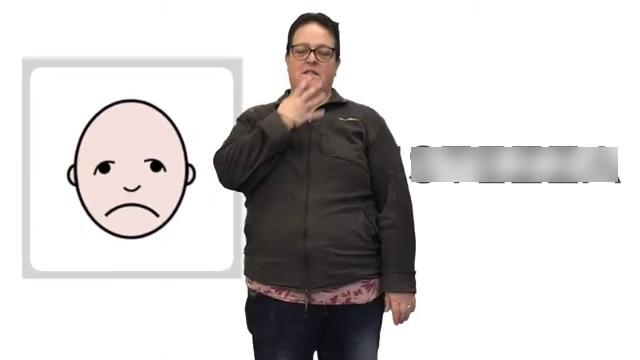} & 
\centering False & 
\begin{tabular}{@{}l@{}}
GT: \textit{emotion\_sadness} \\
Pred: \textit{emotion\_anger}
\end{tabular} \\
\bottomrule
\end{tabular}
\end{table}

\paragraph{Ablation Study}
We conducted an ablation study on various instances of the LLava-OneVision baseline, analyzing the impact of pose information, temporal sampling, and fine-tuning on the SignIT dataset.
\\
\textit{2D Keypoints.} 
We compared two LLaVA-OneVision models in a zero-shot setting: one that takes as input the raw RGB image (LLaVA-OneVision w/o 2D Keypoints), and another that takes the RGB image with 2D keypoints of the hands, face, and body projected onto it (LLaVA-OneVision with 2D Keypoints). Results, reported in Table \ref{tab:llava_inference_performance}, indicate that incorporating pose information improve gesture recognition in terms of F1 score (0.202 vs. 0.158), precision (0.348 vs. 0.208), and recall (0.169 vs. 0.146). In terms of accuracy, the two models obtain similar performance (0.137 vs. 0.138). When analyzing performance, in terms of F1 score, across single macro categories, the inclusion of 2D keypoints proves beneficial for improving F1 scores for \textit{Animals} (0.335 vs. 0.262), \textit{Emotions} (0.295 vs. 0.180) and \textit{Family} (0.189 vs. 0.135). Figure~\ref{tab:pose_comparison}, show a qualitative example highlights how pose information can help the model better distinguish certain gestures.
\\

\begin{table*}[!t]
\centering
\caption{Macro-level performance of LLaVA Only inference models on the testing set across different macro categories.}
\label{tab:llava_inference_performance}
\resizebox{\textwidth}{!}{
\tiny
\setlength{\tabcolsep}{2pt}
\begin{tabular}{@{}lcccc|cccc|cccc|cccc|cccc|cccc@{}}
\toprule
\textbf{Model} & \multicolumn{4}{c|}{\textbf{Macro}} & \multicolumn{4}{c|}{\textbf{Animals}} & \multicolumn{4}{c|}{\textbf{Food}} & \multicolumn{4}{c|}{\textbf{Colors}} & \multicolumn{4}{c|}{\textbf{Emotions}} & \multicolumn{4}{c}{\textbf{Family}} \\
\cmidrule(lr){2-5} \cmidrule(lr){6-9} \cmidrule(lr){10-13} \cmidrule(lr){14-17} \cmidrule(lr){18-21} \cmidrule(lr){22-25}
 & Acc & F1 & Pr & Rc & Acc & F1 & Pr & Rc & Acc & F1 & Pr & Rc & Acc & F1 & Pr & Rc & Acc & F1 & Pr & Rc & Acc & F1 & Pr & Rc \\
\midrule
with 2D Keypoints 
& 0.137 & \textbf{0.202} & \textbf{0.348} & \textbf{0.169} 
& \textbf{0.224} & \textbf{0.335} & \textbf{0.779} & \textbf{0.262} 
& 0.075 & 0.123 & 0.099 & 0.114 
& 0.069 & 0.066 & 0.131 & 0.091 
& \textbf{0.189} & \textbf{0.295} & \textbf{0.439} & \textbf{0.210} 
& \textbf{0.127} & \textbf{0.189} & \textbf{0.293} & \textbf{0.169} \\
w/o 2D Keypoints 
& \textbf{0.138} & 0.158 & 0.208 & 0.146
& 0.211 & 0.262 & 0.337 & 0.244 
& \textbf{0.110} & \textbf{0.133} & \textbf{0.162} & \textbf{0.125} 
& \textbf{0.111} & \textbf{0.080} & \textbf{0.120} & \textbf{0.080} 
& 0.137 & 0.180 & 0.273 & 0.145 
& 0.121 & 0.135 & 0.150 & 0.135 \\
\bottomrule
\end{tabular}
}
\end{table*}

\begin{table*}[!t]
\centering
\caption{Effect of fine-tuning on LLaVA performance across different macro categories (testing set).}
\label{tab:llava_macro_finetuning}
\resizebox{\textwidth}{!}{
\tiny
\setlength{\tabcolsep}{2pt}
\begin{tabular}{@{}lcccc|cccc|cccc|cccc|cccc|cccc@{}}
\toprule
\textbf{Model} & \multicolumn{4}{c|}{\textbf{Macro}} & \multicolumn{4}{c|}{\textbf{Animals}} & \multicolumn{4}{c|}{\textbf{Food}} & \multicolumn{4}{c|}{\textbf{Colors}} & \multicolumn{4}{c|}{\textbf{Emotions}} & \multicolumn{4}{c}{\textbf{Family}} \\
\cmidrule(lr){2-5} \cmidrule(lr){6-9} \cmidrule(lr){10-13} \cmidrule(lr){14-17} \cmidrule(lr){18-21} \cmidrule(lr){22-25}
 & Acc & F1 & Pr & Rc & Acc & F1 & Pr & Rc & Acc & F1 & Pr & Rc & Acc & F1 & Pr & Rc & Acc & F1 & Pr & Rc & Acc & F1 & Pr & Rc \\
\midrule
LLaVA-OneVision Fine-tuned 
& \textbf{0.635} & \textbf{0.644} & \textbf{0.648} & \textbf{0.646}
& \textbf{0.738} & \textbf{0.730} & \textbf{0.712} & \textbf{0.750} 
& \textbf{0.769} & \textbf{0.706} & \textbf{0.643} & \textbf{0.783} 
& \textbf{0.570} & \textbf{0.579} & \textbf{0.579} & \textbf{0.579} 
& \textbf{0.492} & \textbf{0.545} & \textbf{0.600} & \textbf{0.500} 
& \textbf{0.607} & \textbf{0.659} & \textbf{0.707} & \textbf{0.617} \\
LLaVA-OneVision Pre-trained 
& 0.138 & 0.158 & 0.208 & 0.146
& 0.211 & 0.262 & 0.337 & 0.244 
& 0.110 & 0.133 & 0.162 & 0.125 
& 0.111 & 0.080 & 0.120 & 0.080 
& 0.137 & 0.180 & 0.273 & 0.145 
& 0.121 & 0.135 & 0.150 & 0.135 \\
\bottomrule
\end{tabular}
}
\end{table*}

\begin{table*}[!t]
\centering
\caption{Comparison of LLaVA fine-tuned models (Sequence vs. Single Frame) across different macro categories (testing set).}
\label{tab:llava_ft_comparison}
\resizebox{\textwidth}{!}{
\tiny
\setlength{\tabcolsep}{2pt}
\begin{tabular}{@{}lcccc|cccc|cccc|cccc|cccc|cccc@{}}
\toprule
\textbf{Model} & \multicolumn{4}{c|}{\textbf{Macro}} & \multicolumn{4}{c|}{\textbf{Animals}} & \multicolumn{4}{c|}{\textbf{Food}} & \multicolumn{4}{c|}{\textbf{Colors}} & \multicolumn{4}{c|}{\textbf{Emotions}} & \multicolumn{4}{c}{\textbf{Family}} \\
\cmidrule(lr){2-5} \cmidrule(lr){6-9} \cmidrule(lr){10-13} \cmidrule(lr){14-17} \cmidrule(lr){18-21} \cmidrule(lr){22-25}
 & Acc & F1 & Pr & Rc & Acc & F1 & Pr & Rc & Acc & F1 & Pr & Rc & Acc & F1 & Pr & Rc & Acc & F1 & Pr & Rc & Acc & F1 & Pr & Rc \\
\midrule
LLaVA-OneVision (Sequence) 
& 0.424 & 0.318 & 0.489 & 0.423
& 0.364 & 0.471 & 0.667 & 0.364 
& 0.500 & 0.195 & 0.121 & 0.500 
& 0.500 & 0.195 & 0.327 & 0.498
& \textbf{0.571} & 0.421 & 0.333 & \textbf{0.571} 
& 0.182 & 0.308 & \textbf{1.00} & 0.182 \\
LLaVA-OneVision (Single Frame) 
& \textbf{0.635} & \textbf{0.643} & \textbf{0.648} & \textbf{0.646}
& \textbf{0.738} & \textbf{0.730} & \textbf{0.712} & \textbf{0.750} 
& \textbf{0.769} & \textbf{0.706} & \textbf{0.643} & \textbf{0.783}
& \textbf{0.570} & \textbf{0.579} & \textbf{0.579} & \textbf{0.579} 
& 0.492 & \textbf{0.545} & \textbf{0.600} & 0.500 
& \textbf{0.607} & \textbf{0.659} & 0.707 & \textbf{0.617} \\
\bottomrule
\end{tabular}
}
\end{table*}

\textit{Fine-tuning.}

We compare two models that both use human pose information on the frame: one is a fine-tuned model, and the other is a pre-trained model. 

\textit{Temporal Sampling.} 

We compared the effect of temporal sampling on the input of the LLaVA-OneVision model fine-tuned on the SignIT dataset. Specifically, we evaluated two configurations: one where the model receives only the central frame of the video clip as input (i.e., Single Frame), and another where it processes a sequence of four frames uniformly sampled from the entire clip (Sequence). In all experiments, every frame provided to the model includes full body pose information.
Table {\ref{tab:llava_ft_comparison}} presents the results, showing that using only the central frame of the input video is sufficient to classify gestures into one of the five considered macro-categories. In particular, the model that processes only the central frame outperforms the one using a sequence of frames in terms of accuracy (0.635 vs. 0.424), F1 score (0.643 vs. 0.318), precision (0.648 vs. 0.489), and recall (0.646 vs. 0.423). This confirms that the five macro-categories are visually distinct and can be effectively discriminated using a single image.

\subsubsection{Micro-categories}

In this setting, we evaluated the models’ ability to distinguish the 94 individual signs of the SignIT dataset within their respective macro-categories. For each baseline, we trained a model on the classes belonging to the corresponding macro-category. Specifically, \textit{Animals} includes 32 subclasses, \textit{Food} 20 subclasses, \textit{Colors} 17 subclasses, \textit{Emotions} 5 subclasses, and \textit{Family} 20 subclasses.
Table~\ref{tab:micro_overview} shows the results of the considered baselines. The first column reports the evaluation metrics as average across all instances trained for each macro-category.
Results show that the LLaVA-OneVision (last row) outperforms all other baselines, achieving the highest accuracy (0.152), F1-score (0.206), precision (0.359), and recall (0.169). Overall, LLaVA-OneVision consistently surpasses the other baselines across all macro-categories. This instance of LLaVA-OneVision is a fine-tuned model that takes in input a sequence of frames with 2D Keypoints suggesting that to recognize these fine-grained signs, poses and multiple frames are necessary.

In general, all baselines highlight significant difficulties in classifying fine-grained signs within each macro-category, with performance considerably lower than at the macro level highlighting how much is challenging classify the 94 categories of the SignIT dataset.

\begin{table*}[ht!]
\centering

\caption{Micro-level performance across models and categories of the SignIT dataset. Best results are in \textbf{bold}, second best results are \underline{underlined}.}
\label{tab:micro_overview}
\resizebox{\textwidth}{!}{
\tiny
\setlength{\tabcolsep}{2.5pt}
\begin{tabular}{@{}lcccc|cccc|cccc|cccc|cccc|cccc@{}}
\toprule
& \multicolumn{4}{c}{\textbf{Micro}} & \multicolumn{4}{c}{\textbf{Animals}} & \multicolumn{4}{c}{\textbf{Food}} & \multicolumn{4}{c}{\textbf{Colors}} & \multicolumn{4}{c}{\textbf{Emotions}} & \multicolumn{4}{c}{\textbf{Family}} \\
\cmidrule(lr){2-5} \cmidrule(lr){6-9} \cmidrule(lr){10-13} \cmidrule(lr){14-17} \cmidrule(lr){18-21} \cmidrule(lr){22-25}
\textbf{Model} 
& \textbf{Acc} & \textbf{F1} & \textbf{P} & \textbf{R} 
& \textbf{Acc} & \textbf{F1} & \textbf{P} & \textbf{R} 
& \textbf{Acc} & \textbf{F1} & \textbf{P} & \textbf{R} 
& \textbf{Acc} & \textbf{F1} & \textbf{P} & \textbf{R}
& \textbf{Acc} & \textbf{F1} & \textbf{P} & \textbf{R}
& \textbf{Acc} & \textbf{F1} & \textbf{P} & \textbf{R} \\
\midrule
MLP 
& 0.082 & \underline{0.082} & \underline{0.197} & 0.082 
& \underline{0.032} & 0.019 & 0.016 & \underline{0.032}
& 0.005 & 0.004 & 0.009 & 0.005
& 0.027 & 0.021 & 0.022 & 0.027
& \underline{0.201} & 0.222 & 0.291 & 0.201
& 0.145 & 0.152 & 0.158 & 0.145 \\
ResNet-18 
& 0.082 & 0.068 & 0.076 & 0.061
& 0.025 & \underline{0.021} & \underline{0.081} & 0.025
& 0.001 & 0.001 & 0.002 & 0.001
& 0.036 & 0.037 & 0.024 & 0.036
& 0.149 & 0.111 & 0.106 & 0.121
& \textbf{0.171} & \underline{0.168} & \underline{0.177} & \textbf{0.171} \\
I3D 
& \underline{0.124} & 0.046 & 0.041 & 0.073
& 0.023 & 0.003 & 0.002 & 0.023
& \underline{0.044} & 0.003 & 0.002 & \underline{0.044}
& \underline{0.095} & 0.016 & 0.009 & \textbf{0.086}
& 0.121 & 0.111 & 0.100 & 0.121
& 0.120 & 0.093 & 0.089 & 0.120 \\
KNN 
& 0.089 & 0.130 & 0.132 & \underline{0.089}
& 0.021 & 0.017 & 0.017 & 0.021
& 0.017 & \underline{0.012} & \underline{0.012} & 0.017
& 0.042 & \underline{0.039} & \underline{0.041} & 0.042
& \textbf{0.270} & \textbf{0.440} & \textbf{0.444} & \textbf{0.270}
& 0.093 & 0.137 & 0.124 & 0.093 \\

\textbf{LLaVA-OneVision}
& \textbf{0.152} & \textbf{0.206} & \textbf{0.359} & \textbf{0.169}
& \textbf{0.224} & \textbf{0.335} & \textbf{0.779} & \textbf{0.262}
& \textbf{0.110} & \textbf{0.133} & \textbf{0.162} & \textbf{0.125}
& \textbf{0.111} & \textbf{0.080} & \textbf{0.120} & \underline{0.080}
& 0.189 & \underline{0.295} & \underline{0.439} & \underline{0.210}
& \underline{0.127} & \textbf{0.189} & \textbf{0.293} & \underline{0.169} \\
\bottomrule
\end{tabular}
}

\end{table*}

\begin{figure}[!t]
\centering
\scriptsize
\setlength{\tabcolsep}{1pt}
\resizebox{\columnwidth}{!}{%
\renewcommand{\arraystretch}{1.2}
\begin{tabular}{@{}cl@{}}

\begin{tabular}{@{}ccc@{}}
\includegraphics[width=0.095\textwidth,height=0.127\textwidth,keepaspectratio=false]{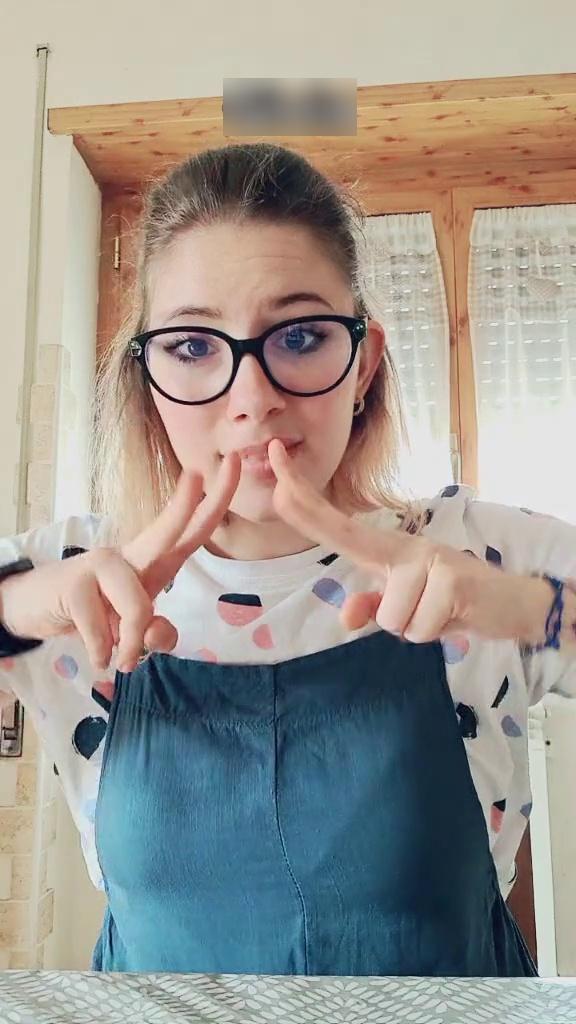} &
\includegraphics[width=0.095\textwidth,height=0.127\textwidth,keepaspectratio=false]{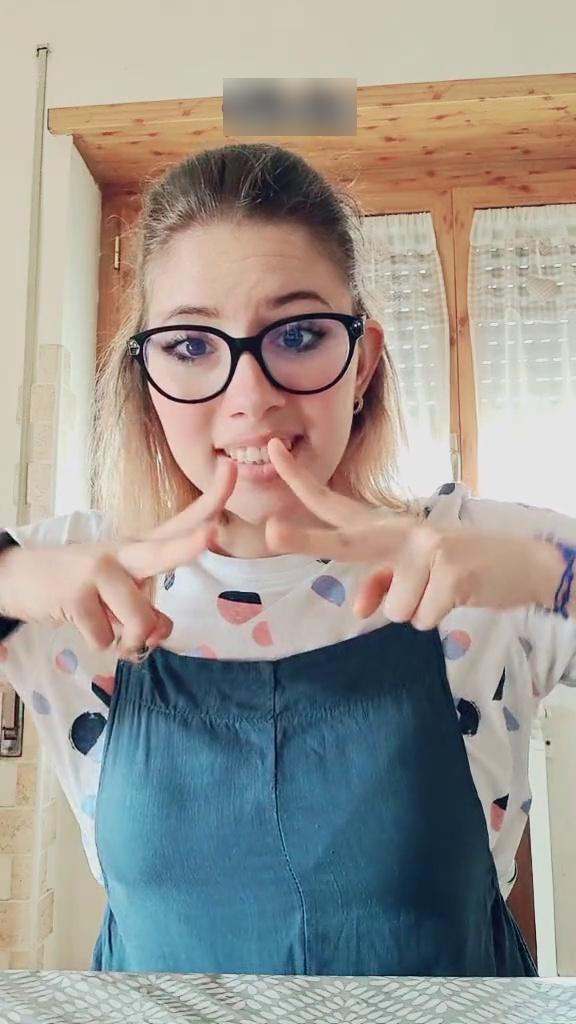} &
\includegraphics[width=0.095\textwidth,height=0.127\textwidth,keepaspectratio=false]{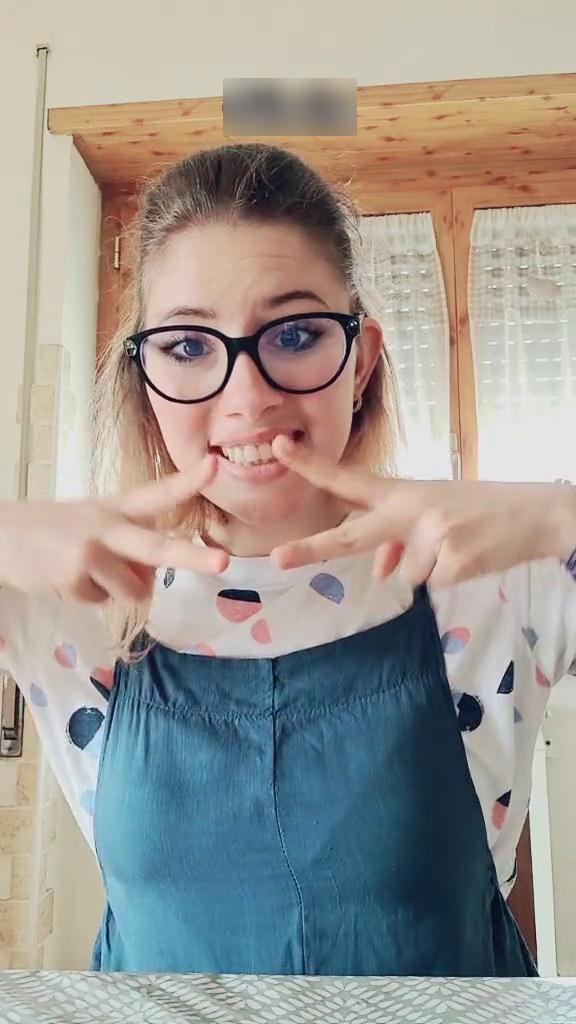}
\end{tabular}
&
\begin{tabular}{@{}ll@{}}
\textbf{KNN:} & \textcolor{red}{disgust} \\
\textbf{MLP:} & \textcolor{red}{fear} \\
\textbf{ResNet:} & \textcolor{red}{anger} \\
\textbf{I3D:} & \textcolor{green!60!black}{joy} \\
\textbf{LLaVA:} & \textcolor{green!60!black}{joy}
\end{tabular} \\[4pt]
\begin{tabular}{@{}ccc@{}}
\includegraphics[width=0.095\textwidth,height=0.127\textwidth,keepaspectratio=false]{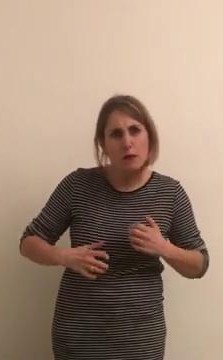} &
\includegraphics[width=0.095\textwidth,height=0.127\textwidth,keepaspectratio=false]{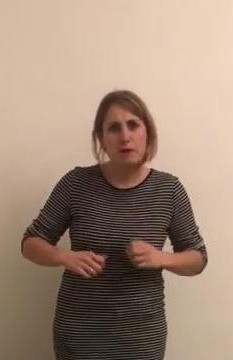} &
\includegraphics[width=0.095\textwidth,height=0.127\textwidth,keepaspectratio=false]{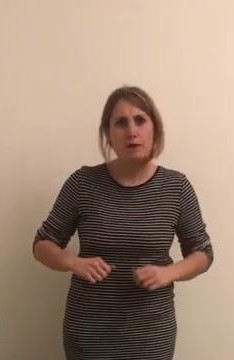}
\end{tabular}
&
\begin{tabular}{@{}ll@{}}
\textbf{KNN:} & \textcolor{red}{sadness} \\
\textbf{MLP:} & \textcolor{green!60!black}{anger} \\
\textbf{ResNet:} & \textcolor{green!60!black}{anger} \\
\textbf{I3D:} & \textcolor{red}{sadness} \\
\textbf{LLaVA:} & \textcolor{green!60!black}{anger}
\end{tabular} \\[4pt]
\begin{tabular}{@{}ccc@{}}
\includegraphics[width=0.095\textwidth,height=0.127\textwidth,keepaspectratio=false]{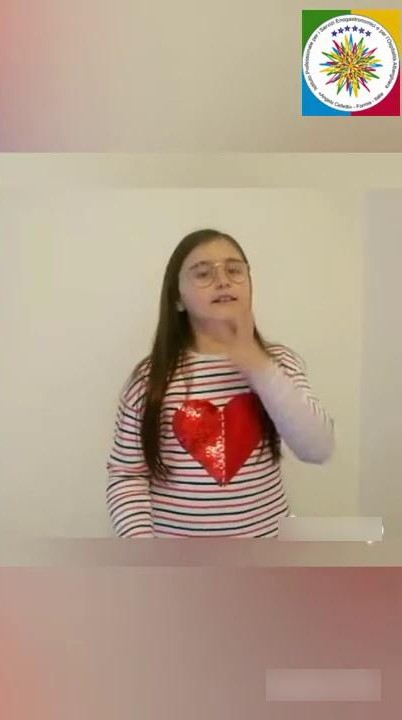} &
\includegraphics[width=0.095\textwidth,height=0.127\textwidth,keepaspectratio=false]{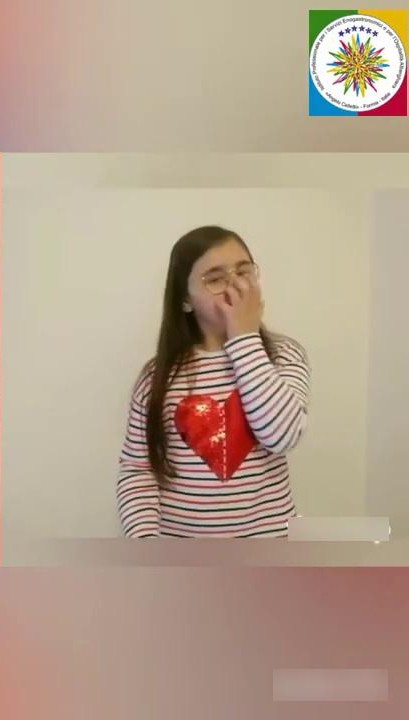} &
\includegraphics[width=0.095\textwidth,height=0.127\textwidth,keepaspectratio=false]{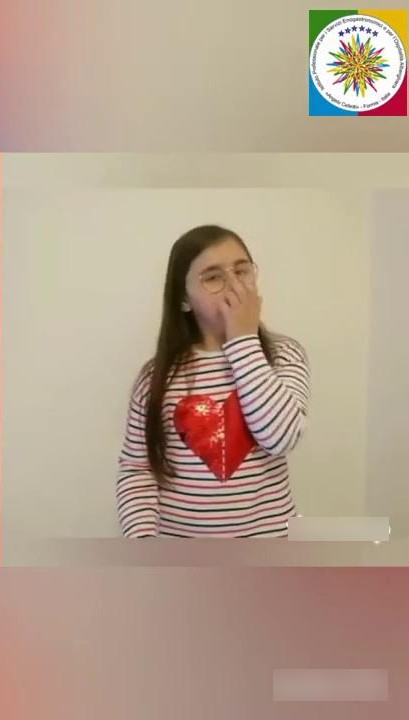}
\end{tabular}
&
\begin{tabular}{@{}ll@{}}
\textbf{KNN:} & \textcolor{green!60!black}{sadness} \\
\textbf{MLP:} & \textcolor{red}{disgust} \\
\textbf{ResNet:} & \textcolor{red}{disgust} \\
\textbf{I3D:} & \textcolor{red}{joy} \\
\textbf{LLaVA:} & \textcolor{red}{fear}
\end{tabular} \\

\end{tabular}
}
\caption{Model predictions comparison on sample frames. Green: correct, red: incorrect.}
\label{fig:model_comparison}
\end{figure}
\begin{table*}[!t]
\centering
\caption{Micro-level performance of LLaVA fine-tuned models with different temporal sampling strategies across macro categories.}
\label{tab:llava_ft_micro_extended}
\resizebox{\textwidth}{!}{
\tiny
\setlength{\tabcolsep}{2pt}
\begin{tabular}{@{}lcccc|cccc|cccc|cccc|cccc|cccc@{}}
\toprule
\textbf{Model} &
\multicolumn{4}{c|}{\textbf{Average}} &
\multicolumn{4}{c|}{\textbf{Animals}} &
\multicolumn{4}{c|}{\textbf{Food}} &
\multicolumn{4}{c|}{\textbf{Colors}} &
\multicolumn{4}{c|}{\textbf{Emotions}} &
\multicolumn{4}{c}{\textbf{Family}} \\
\cmidrule(lr){2-5} \cmidrule(lr){6-9} \cmidrule(lr){10-13}
\cmidrule(lr){14-17} \cmidrule(lr){18-21} \cmidrule(lr){22-25}
& Acc & F1 & Pr & Rc
& Acc & F1 & Pr & Rc
& Acc & F1 & Pr & Rc
& Acc & F1 & Pr & Rc
& Acc & F1 & Pr & Rc
& Acc & F1 & Pr & Rc \\
\midrule

LLaVA-OneVision (Sequence)
& \textbf{0.238} & \textbf{0.192} & \textbf{0.206} & \textbf{0.209}
& \textbf{0.510} & \textbf{0.372} & \textbf{0.391} & \textbf{0.429} 
& \textbf{0.167} & \textbf{0.142} & \textbf{0.158} & \textbf{0.150}
& \textbf{0.124} & \textbf{0.110} & \textbf{0.118} & \textbf{0.115}
& \textbf{0.201} & \textbf{0.176} & \textbf{0.188} & \textbf{0.183}
& \textbf{0.188} & \textbf{0.160} & \textbf{0.175} & \textbf{0.168} \\

LLaVA-OneVision (Single Frame)
& 0.138 & 0.102 & 0.140 & 0.119 
& 0.168 & 0.122 & 0.166 & 0.149
& 0.131 & 0.098 & 0.142 & 0.113
& 0.093 & 0.070 & 0.083 & 0.076
& 0.154 & 0.118 & 0.162 & 0.139
& 0.142 & 0.101 & 0.147 & 0.119 \\

\bottomrule
\end{tabular}
}
\end{table*}

\begin{table*}[!t]
\centering
\caption{Effect of fine-tuning on LLaVA performance across macro categories for micro-level classification (testing set).}
\label{tab:llava_micro_finetuning_extended}
\resizebox{\textwidth}{!}{
\tiny
\setlength{\tabcolsep}{2pt}
\begin{tabular}{@{}lcccc|cccc|cccc|cccc|cccc|cccc@{}}
\toprule
\textbf{Model} &
\multicolumn{4}{c|}{\textbf{Average}} &
\multicolumn{4}{c|}{\textbf{Animals}} &
\multicolumn{4}{c|}{\textbf{Food}} &
\multicolumn{4}{c|}{\textbf{Colors}} &
\multicolumn{4}{c|}{\textbf{Emotions}} &
\multicolumn{4}{c}{\textbf{Family}} \\
\cmidrule(lr){2-5} \cmidrule(lr){6-9} \cmidrule(lr){10-13}
\cmidrule(lr){14-17} \cmidrule(lr){18-21} \cmidrule(lr){22-25}
& Acc & F1 & Pr & Rc
& Acc & F1 & Pr & Rc
& Acc & F1 & Pr & Rc
& Acc & F1 & Pr & Rc
& Acc & F1 & Pr & Rc
& Acc & F1 & Pr & Rc \\
\midrule

LLaVA-OneVision (Fine-tuned with Pose)
& \textbf{0.238} & \textbf{0.192} & \textbf{0.206} & \textbf{0.209}
& \textbf{0.510} & \textbf{0.378} & \textbf{0.392} & \textbf{0.432} 
& \textbf{0.167} & \textbf{0.141} & \textbf{0.158} & \textbf{0.149}
& \textbf{0.124} & \textbf{0.108} & \textbf{0.118} & \textbf{0.114}
& \textbf{0.201} & \textbf{0.174} & \textbf{0.188} & \textbf{0.183}
& \textbf{0.188} & \textbf{0.159} & \textbf{0.174} & \textbf{0.167} \\

LLaVA-OneVision Pre-Trained (Pose)
& 0.121 & 0.135 & 0.150 & 0.135
& 0.244 & 0.282 & 0.280 & 0.275 
& 0.085 & 0.094 & 0.108 & 0.097
& 0.060 & 0.072 & 0.083 & 0.069
& 0.114 & 0.119 & 0.142 & 0.121
& 0.102 & 0.108 & 0.137 & 0.113 \\

LLaVA-OneVision Pre-Trained (No-Pose)
& \underline{0.127} & \underline{0.189} & \underline{0.293} & \underline{0.169}
& \underline{0.223} & \underline{0.366} & \underline{0.485} & \underline{0.350} 
& \underline{0.094} & \underline{0.132} & \underline{0.226} & \underline{0.115}
& \underline{0.071} & \underline{0.097} & \underline{0.181} & \underline{0.082}
& \underline{0.128} & \underline{0.183} & \underline{0.309} & \underline{0.156}
& \underline{0.119} & \underline{0.167} & \underline{0.264} & \underline{0.142} \\

\bottomrule
\end{tabular}
}
\end{table*}

\paragraph{Ablation Study}
We conducted a comprehensive ablation study similar to the one conducted for the macro categories setting. Specifically, we analyze the impact of temporal sampling strategy, fine-tuning, and explicit pose cues on micro-level classification.

\textit{Temporal Sampling.}
Table~\ref{tab:llava_ft_comparison} compares the performance of fine-tuned models utilizing different temporal sampling strategies. Unlike the macro-level classification, where processing a single central frame proved sufficient due to the coarse nature of the distinction, temporal information plays a crucial role at the micro level. The sequence-based model (\textbf{LLaVA FT (Sequence w/ Pose)}) significantly outperforms the single-frame variant (\textbf{LLaVA FT (Single Frame w/ Pose)}) across all metrics. Specifically, accuracy improves from $0.138$ to $\mathbf{0.238}$ ($+72,96\%$ relative gain), and the F1-score increases substantially from $0.102$ to $\mathbf{0.192}$ ($+84,24\%$ relative gain). This demonstrates that distinguishing between visually similar signs within the same semantic macro-category requires capturing temporal dynamics and movement patterns. For fine-grained recognition, the kinetic features encoded by the sequence model are essential discriminative elements that a static image cannot resolve.

\smallskip

\textit{Fine-tuning and 2D Keypoints.}
We compared three instances of the LLaVA-OneVision model that take as input a sequence of frames. The first instance was fine-tuned and leverages 2D keypoints (LLaVA-OneVision Fine-tuned with Pose), the second is a pre-trained model that also uses pose information, and the third is a pre-trained model that relies solely on RGB frames.
Table~\ref{tab:llava_micro_finetuning_extended}) shows that domain-specific adaptation is essential for micro-level sign recognition. The fine-tuned model (first row) achieves best results for all the evaluation metrics. Specifically, it obtains an accuracy of $0.238$, an F1-score of $0.192$, a precision of $0.206$ and a recall of $0.209$ demonstrating that fine-tuning the model and using 2D Keypoints is beneficial for the recognition of the 94 signs of the SignIT dataset.

\section{\uppercase{Implementation Details}}
\label{sec:implementation}

To establish robust and reproducible performance benchmarks for the SignIT dataset, we implemented a comprehensive set of baseline models across three distinct input modalities: 2D human pose keypoints, visual appearance (RGB frames), and a multimodal combination of both. The entire framework was built upon the \textbf{PyTorch} deep learning library.

The input data underwent stringent preprocessing. Keypoint features, extracted by MediaPipe for hands, face, and body, were first concatenated into a $\mathbf{210}$-dimensional vector per frame. These features were subsequently normalized using \textbf{Z-score standardization} based on the training set statistics, ensuring invariance to scaling and absolute position. For the lightweight keypoint-based models, \textbf{K-Nearest Neighbors (K-NN)} utilized a Euclidean distance metric with $K=5$, and the \textbf{Multi-Layer Perceptron (MLP)} employed a three-layer architecture (Input-512-Output) trained for $\mathbf{30}$ epochs using \textbf{AdamW} ($lr=1\times10^{-4}$) and a time-averaged keypoint vector as input.

Visual appearance models were initialized with pre-trained weights to leverage prior knowledge. \textbf{ResNet-18} was pre-trained on ImageNet and required only a single center-cropped $224 \times 224$ frame for classification. The spatio-temporal dynamics were modeled by \textbf{I3D (Inflated 3D ConvNet)}, pre-trained on Kinetics-400, which processed input clips consisting of $\mathbf{16}$ uniformly sampled frames. Both models were trained with \textbf{Cross-Entropy Loss}.
To adapt the \textbf{LLaVA-OneVision-Qwen2-7B} general-purpose model to the fine-grained specifics of LIS gestures, we utilized the parameter-efficient \textbf{LoRA} (Low-Rank Adaptation) method. The LoRA configuration involved a rank of $\mathbf{r=16}$, $\mathbf{\alpha=32}$, and a dropout of $\mathbf{0.05}$. Training was conducted for $\mathbf{10}$ epochs using the \textbf{Paged AdamW 8-bit} optimizer ($lr=2\times10^{-5}$) with FP16 mixed precision, achieving an effective batch size of 16 through gradient accumulation. .

\section{\uppercase{Conclusions}}
\label{sec:conclusion}
We presented the SignIT dataset which comprises sign belonging to the Italian Sign Language (LIS).  It comprises 644 videos with ~99,000 annotated frames across 94 gesture classes in five semantic categories. The dataset integrates hand, face, and body keypoints with RGB frames, enabling comprehensive evaluation of diverse recognition paradigms.
With the dataset, we presented a benchmark for systematically studying the task of sign recognition. We adopted several state-of-the-art baselines and considered different settings (macro and micro categories). The benchmark shows how RGB, 2D Keypoints and temporal information can influence performance of models for addressing this challenging task on the SignIT dataset.
Our work establishes a reproducible resource for LIS recognition research and demonstrates multimodal learning's potential in bridging communication barriers.

\section{\uppercase{Acknowledgements}}
This study has been supported by Next Vision s.r.l. and by the Research Program PIAno di inCEntivi per la Ricerca di Ateneo 2024/2026, project "Multi-Agent Simulator for Real-Time Decision-Making Strategies in Uncertain Egocentric Scenarios" - University of Catania.
\bibliographystyle{apalike}
{\small
\bibliography{References}}

\end{document}